\documentclass[conference]{IEEEtran}

\title{A Customisable Underwater Robot}

\ifCLASSINFOpdf
   \usepackage[pdftex]{graphicx}
   \graphicspath{{./figures/}}
   \DeclareGraphicsExtensions{.pdf,.jpeg,.png}
\else
  	\usepackage[dvips]{graphicx}
   \graphicspath{{./figures/}}
   \DeclareGraphicsExtensions{.eps}
\fi

\author{
\IEEEauthorblockN{Guido Schillaci}
\IEEEauthorblockA{Adaptive Systems Group\\
Computer Science Department\\
Humboldt-Universit\"{a}t zu Berlin\\
guido.schillaci@informatik.hu-berlin.de}\\   
\and
\IEEEauthorblockN{Fabio Schillaci}
\IEEEauthorblockA{Digital Designer and Architect\\
http://www.schillaci.org\\
fabio@schillaci.org}\\
\and
\IEEEauthorblockN{Verena V. Hafner}
\IEEEauthorblockA{Adaptive Systems Group\\
Computer Science Department\\
Humboldt-Universit\"{a}t zu Berlin\\
hafner@informatik.hu-berlin.de}\\
}

\usepackage{gensymb}
\usepackage{hyperref}

\begin{document}

\maketitle

\begin{abstract}
We present a model of a configurable underwater drone, whose parts are optimised for 3D printing processes. We show how - through the use of printable adapters - several thrusters and ballast configurations can be implemented, allowing different maneuvering possibilities.

After introducing the model and illustrating a set of possible configurations, we present a functional prototype based on open source hardware and software solutions. The prototype has been successfully tested in several dives in rivers and lakes around Berlin. 

The reliability of the printed models has been tested only in relatively shallow waters. However, we strongly believe that their availability as freely downloadable models will motivate the general public to build and to test underwater drones, thus speeding up the development of innovative solutions and applications. 

The models and their documentation will be available for download at the following link: 

\url{https://adapt.informatik.hu-berlin.de/schillaci/underwater.html}.

\end{abstract}

\section{Introduction}
Underwater robots are tipically classified into two categories: Autonomous Underwater Vehicles (AUVs) and Remotely Operated Vehicles (ROVs). AUVs are unmanned, self-propelled vehicles that can operate autonomously of a host vessel, which makes them well suited to exploration of extreme environments \cite{Wynn2014}. A continuous development of new vehicles and sensors, and advances in artificial intelligence are increasing the range of AUVs applications \cite{Leonard2016}. However, their high cost and the know-how required to develop them makes their adoption prohibitive to the general public.

Remotely operated underwater vehicles are mobile devices commonly used in deep water industries. ROVs are usually connected to a surface vessel through a tether and controlled by an operator, which makes them more reliable and less prone to failures than AUVs. Similarly to AUVs, professional ROVs are tipically very expensive. 

However, thanks to the rapid growth of opensource software and hardware communities, and thanks to their experience in the development of flying drones, underwater teleoperated drone technology has recently become accessible to a wider public. 
In fact, low-cost open-source solutions for remotely operated vehicles (ROVs) can be found in the market since almost a decade, such as OpenROV (opensource remotely operated vehicle \footnote{\url{https://www.openrov.com/}}) and BlueRov2 (an opensource ROV kit produced by BlueRobotics \footnote{ \url{https://www.bluerobotics.com/}}).

Not surprisingly, research institutes and do-it-yourself communities exhibited a growing interest in the adoption of opensource underwater drones for many applications \cite{siciliano2016springer}: from underwater exploration to marine science, archeology and marine geoscience; from diving to fishing; from boating to harbour and boat maintenance and support; from garbage collection to operation in hazardous environment.

\begin{figure}[htp!]
  \centering
  \includegraphics[scale=0.22]{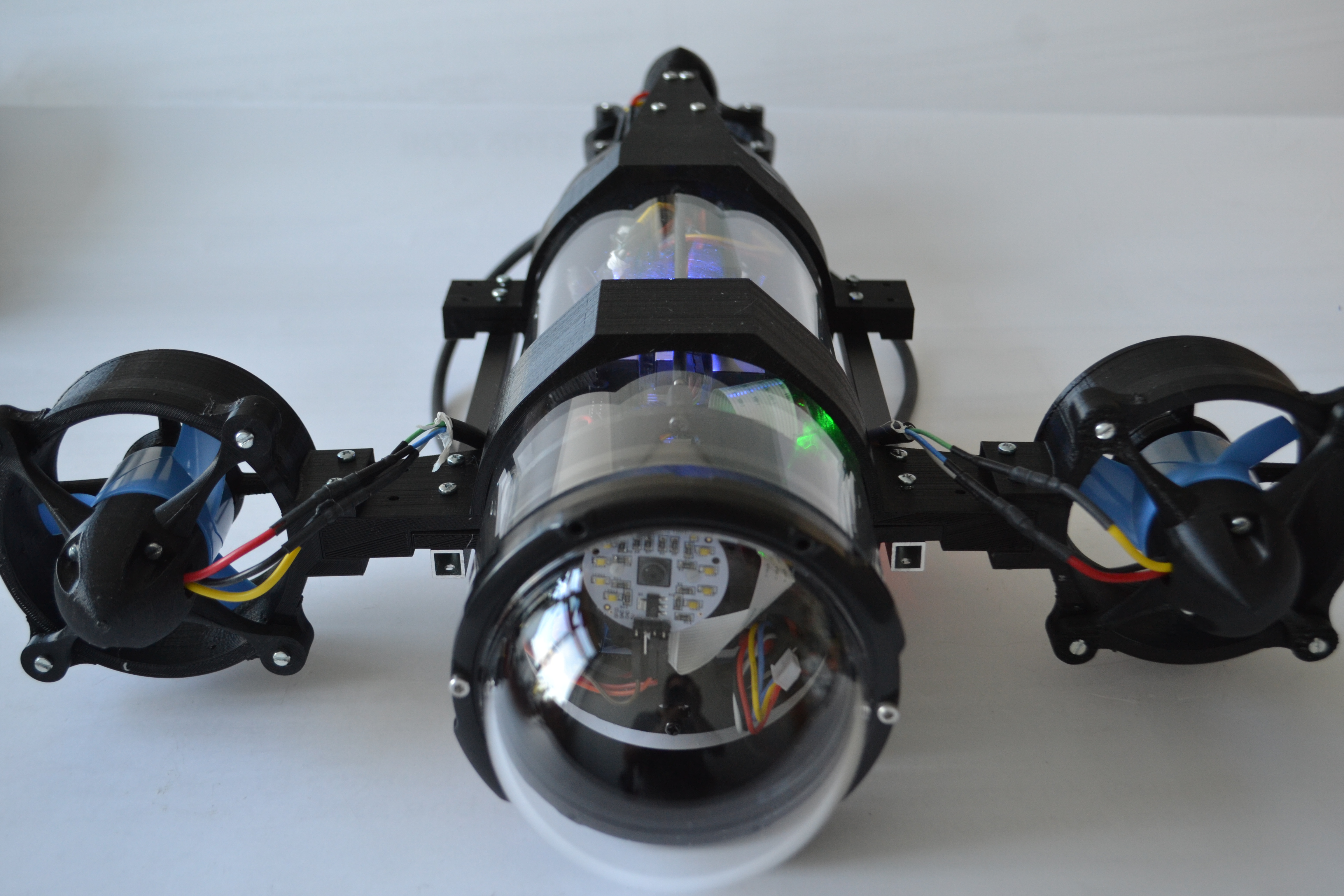}
  \caption{A screenshot of a 3-thrusters prototype.}
  \label{fig:pic}
\end{figure}

Similarly to flying drones, low cost underwater vehicles are usually equipped with cameras and basic sensing capabilities (e.g. inertial measurement unit, depth sensor, etc.), and are teleoperated from the surface. 
Underwater propulsion is achieved through propellers, whose number and configuration (e.g. with fixed or changing orientation) define the degrees of freedom of the movements that the drone can perform underwater.
A conventional procedure in ROVs/AUVs design is to make vehicles positively buyoant to ensure they will return to the surface, should any power failure occurs. Moreover, low-cost ROVs adopt fixed ballast - achieved through lead, buoyancy chambers and syntactic foam - to have the desired gravity. Under a fixed ballast configuration, ascend and descend movements are achieved using vertical thrusters. Variable ballast can surely represent a better solution, as it does not require thusting upwards/downwards, however at the expense of higher cost/complexity of the vehicle.

The electronics of underwater vehicles are enclosed into watertight containers that prevent water infiltration and, consequently, damage to the components. 
Propellers are usually connected to the watertight enclosure through cables and cable glands to ensure watertight connection to the inner electronics.
To reduce the risks of water infiltration due to high water pressure when using cable glands, other solutions have been tested, such as using magnetic coupling to move the propellers using motors positioned inside the watertight enclosure \cite{Hanff2017}.

The components required to build watertight enclosures and connections - together with the electronic components - usually represent a consistent portion of the total cost of the ROV/AUV equipment. Nonetheless, designing and building external components - which includes, for instance, propellers, ballast and waterdynamic configurations - is not a minor, nor cheap task in underwater drones development.

In this paper, we propose an opensource solution for rapid prototyping and customisation of underwater drones. In section \ref{sec:model}, we present 3D-printable models that are compatible with hardware components available in the low-cost underwater drones market (i.e. BlueRobotics and OpenROV) and that can be used to easily customise propellers, ballast and waterdynamic configurations, thus reducing the total cost of the final product.
Therefore, we present an implementation of a functional prototype (section \ref{sec:proto})
that uses the proposed models. Finally, we describe the hardware (section \ref{sec:hw}) and the software solutions (section \ref{sec:sw}) that we adopted in the functional prototype.

\section{A customisable underwater drone}
\label{sec:model}
As of today, several models of ROV parts can be found online, for instance those provided by BlueRobotics. Unfortunately, such models are usually either not optimal for 3D-printing (e.g. walls around screwing holes are thin and thus prone to breakage when mounting the parts) or not modular enough to easily customise the drone configuration without having to buy additional components.

In this work, we present a model for a customisable underwater robot for marine research that addresses these issues. The 3D-models presented here are realeased under an open source license and are available for download at the following link: \url{https://adapt.informatik.hu-berlin.de/schillaci/underwater.html}.
In the development of these models, we took inspiration from the T100/T200 thruster models and on the enclosure clamp model for watertight tubes developed by BlueRobotics \footnote{The original models from BlueRobotics are available at the following link: \url{https://github.com/bluerobotics/bluerobotics.github.io}.}.

The original thruster consists of three plastic parts, plus a propeller to be mounted on the motor. A mounting bracket is provided as a separate model. Initially, we tested printing such models using an Ultimaker 2 3D printer. However, the models are unfortunately not optimal for rapid prototyping techniques.
In fact, we could not find any printer configuration - including solid infill and high quality 0.06 mm layer height - that prevented damages of the plastic while mounting them together using screws, as walls around holes are too thin. It is very likely that such models have been developed for being produced using other techniques, such as injection molding, rather than rapid prototyping through 3D-printers. Moreover, holes in 3D-printed parts can become stripped very quickly, already after a few times bolts are screwed in and unscrewed out. Mounting parts together using nuts and bolts would prevent this, but this option was not possible with the original models provided by BlueRobotics.

\begin{figure}[htp!]
  \centering
  \includegraphics[scale=0.20]{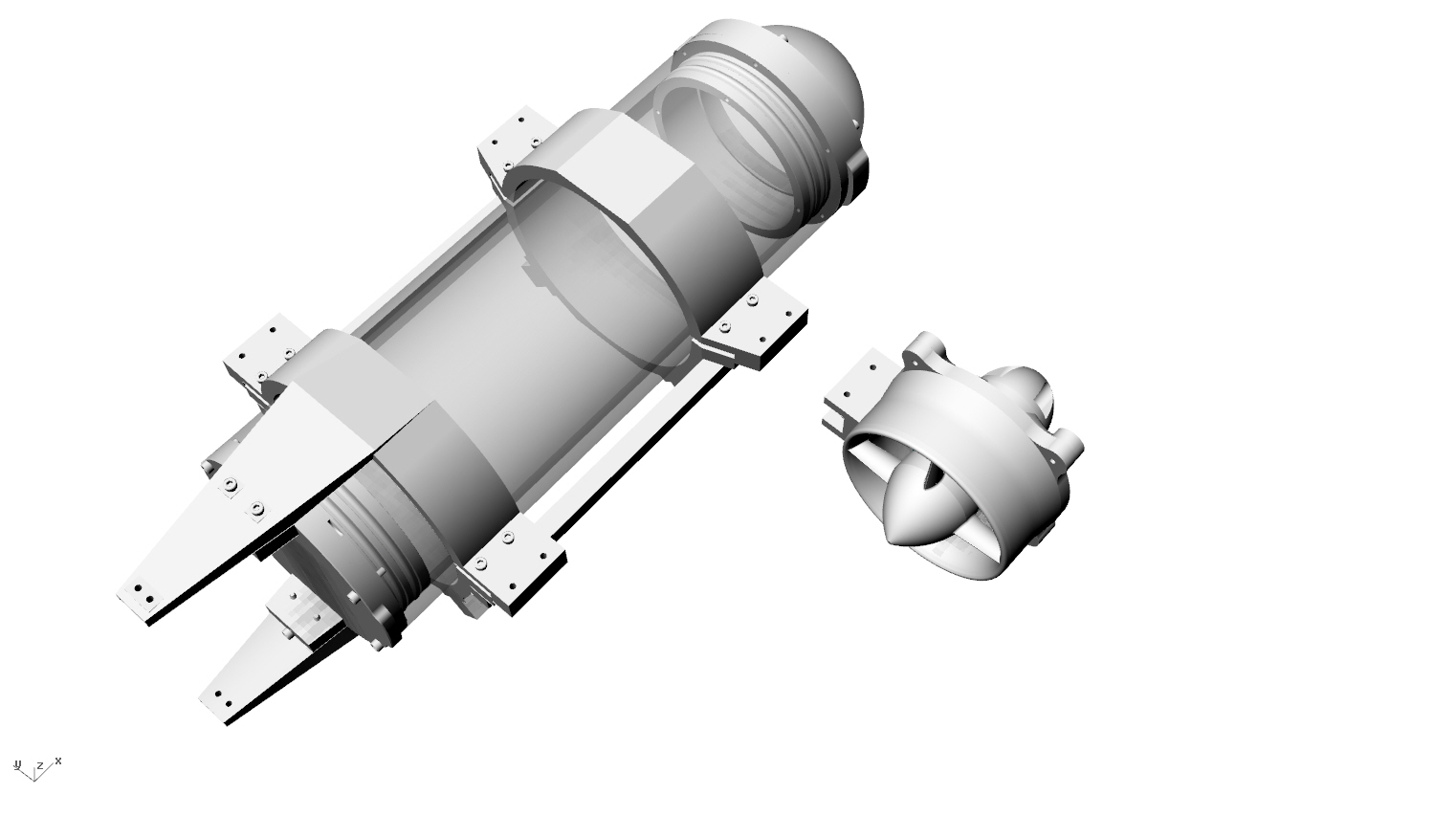}
  \caption{The enclosure clamp is provided with 4 mounting points (+90\degree , +135\degree and -90\degree , -135\degree. This illustration shows an exploded model of a thruster to be mounted on the +90\degree point.}
  \label{fig:mount1}
\end{figure}

Therefore, we re-designed the models - in particular the ones composing the thruster and the enclosure clamp - and optimised them for 3D printing techniques. In particular, we increased the thickness of most of the model parts and we introduced mounting structures that allow using nuts and bolts. We designed an adapter for mounting a cheap Turnigy DST-700 brushless motor (priced in several online stores for \textless 10 EUR) onto the Bluerobotics T200 propeller, which has been specifically designed for the more expensive (priced ca. 80 EUR), although more resistant to corrosion, M200 motor. The new design of the thruster uses a different configuration of the holes that allow easy fastening using M2.5 and M3 nuts and bolts. Moreover, we added mounting points onto the thruster and onto the enclosure clamp models, so that they can be easily mounted together at different positions, as illustrated in Figure \ref{fig:mount1}.

\begin{figure}[ht!]
  \centering
  \includegraphics[scale=0.20]{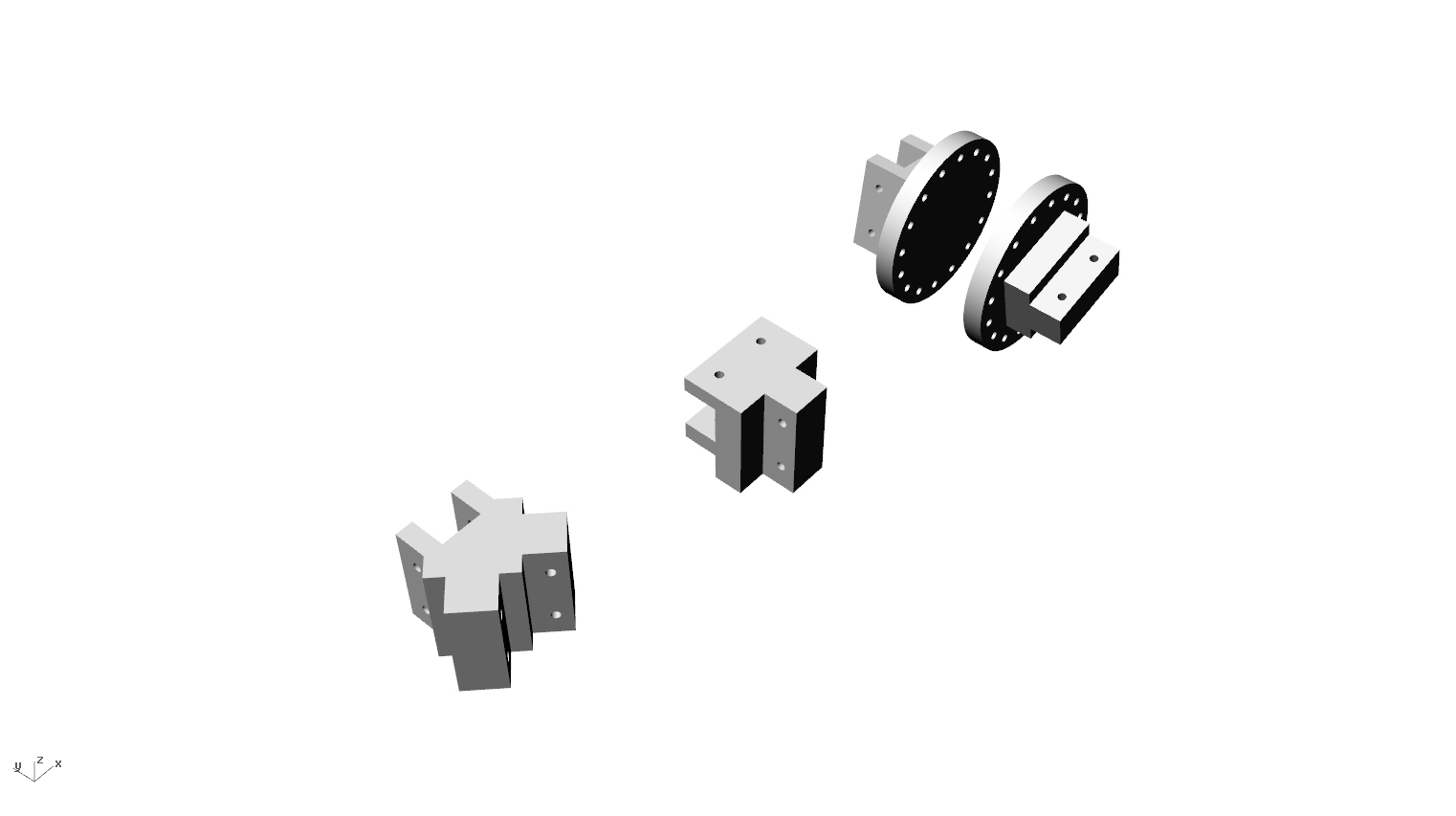}
  \caption{The three adapters that allow setting different orientations of the thursters and ballast.}
  \label{fig:adapters}
\end{figure}

\begin{figure}[htp!]
  \centering
  \includegraphics[scale=0.20]{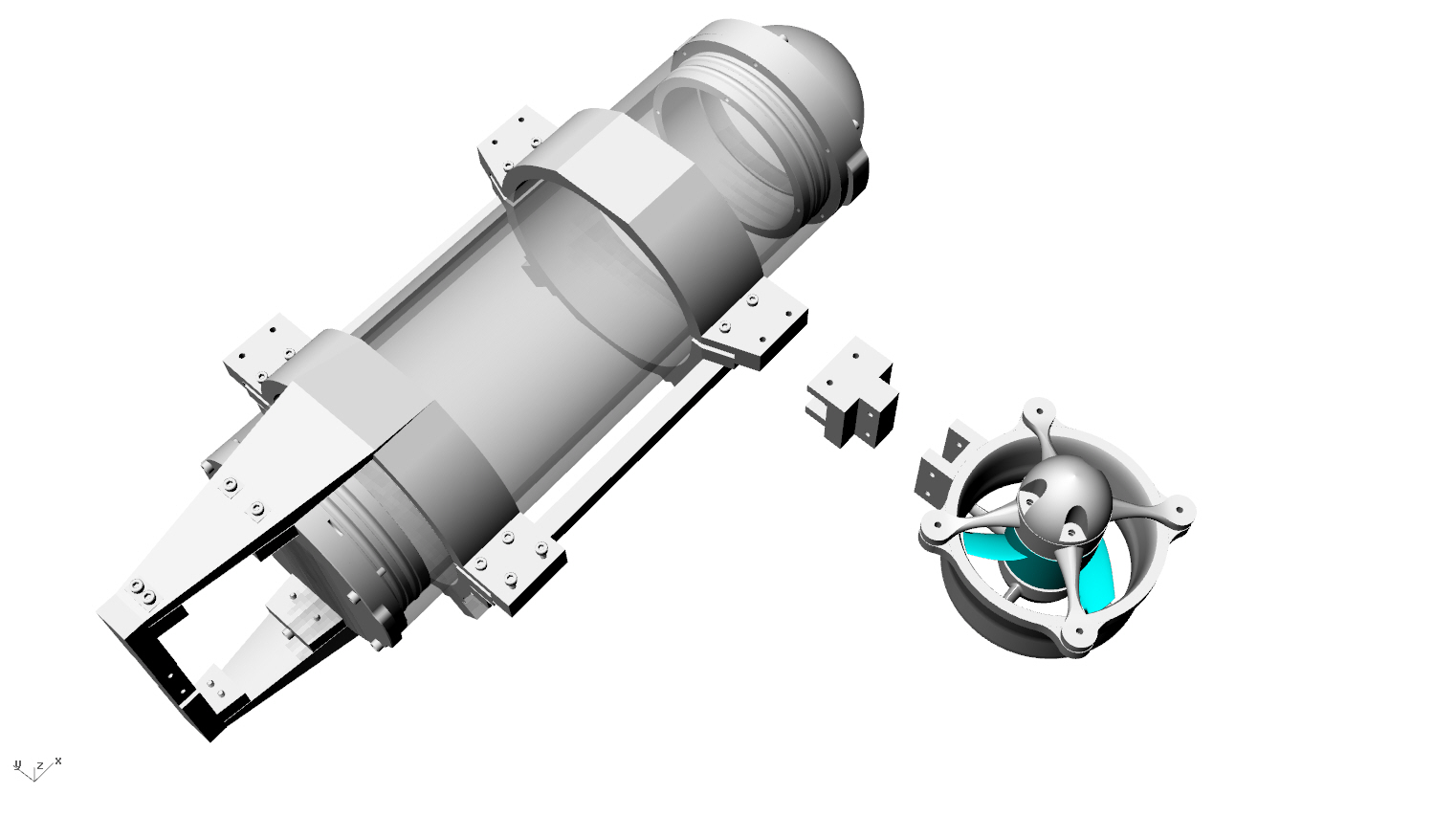}
  \caption{The 90\degree adapter can be used to mount a vertical thurster onto the enclosure clamp.}
  \label{fig:mount2}
\end{figure}

We designed also three adapters that allow to change the orientation of a thruster or to split a mounting point into two, as depicted in Figure \ref{fig:adapters}. For instance, Figure \ref{fig:mount2} shows a thurster to be mounted in a vertical configuration using a 90\degree  adapter. 

\begin{figure}[htp!]
  \centering
  \includegraphics[scale=0.17]{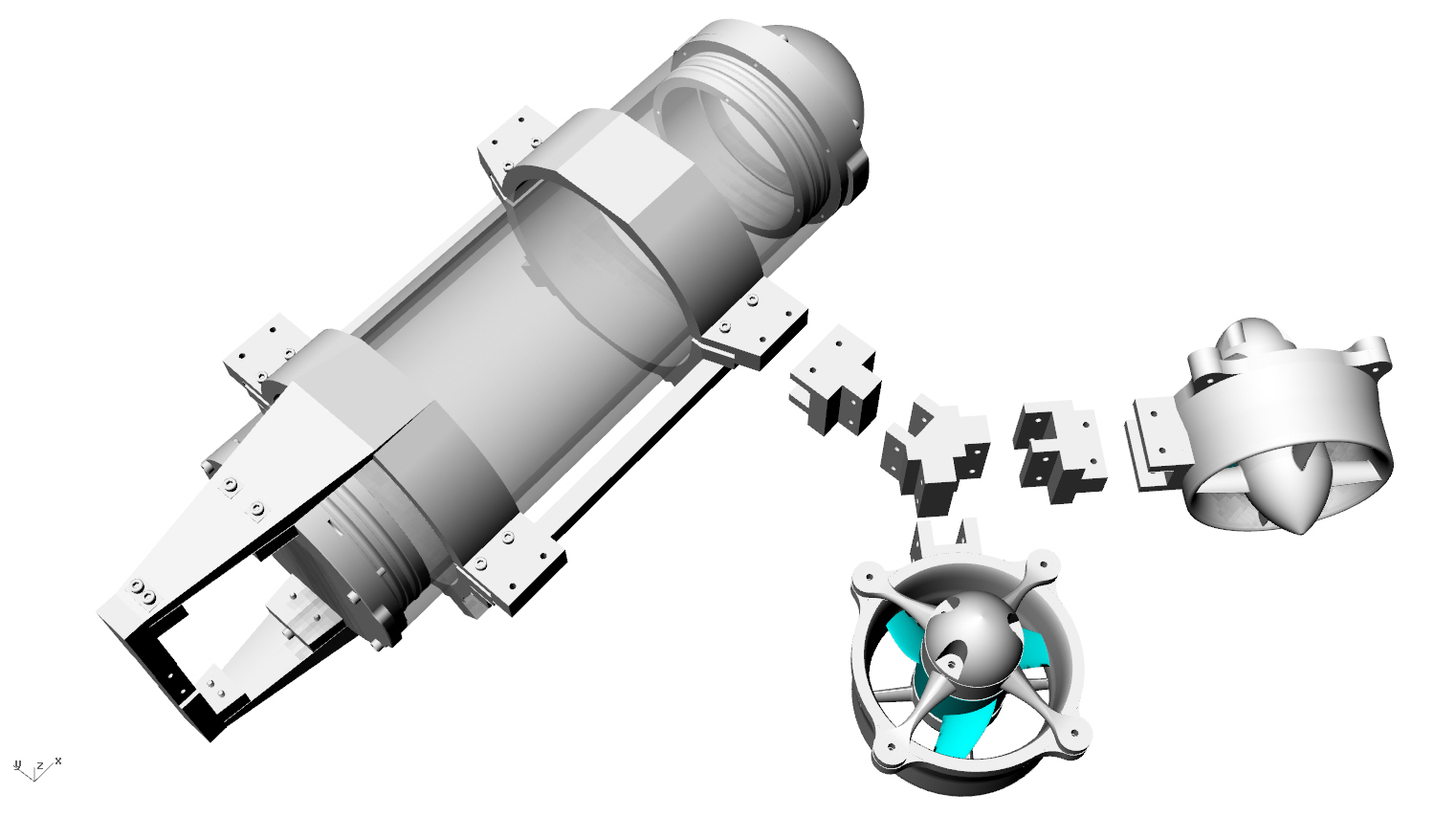}
  \caption{The split adapter can be used to split a mounting point into two. In the figure, a combination of split and 90\degree adapters is used to mount a vertical and a 45\degree horizontal thrusters onto the same mounting point in the enclosure clamp.}
  \label{fig:mount3}
\end{figure}

The split adapter can significantly increase the number of possible configurations, if adopted in combination with the 90\degree  adapter. Figure \ref{fig:mount3} shows a combination of three adapters for mounting a vertical thruster and a 45\degree horizontal thurster onto the same mounting point at the enclosure clamp. Moreover, we designed a rotation adapter that allow five different orientations of a thruster within 90\degree.

\begin{figure}[htp!]
  \centering
  \includegraphics[scale=0.19]{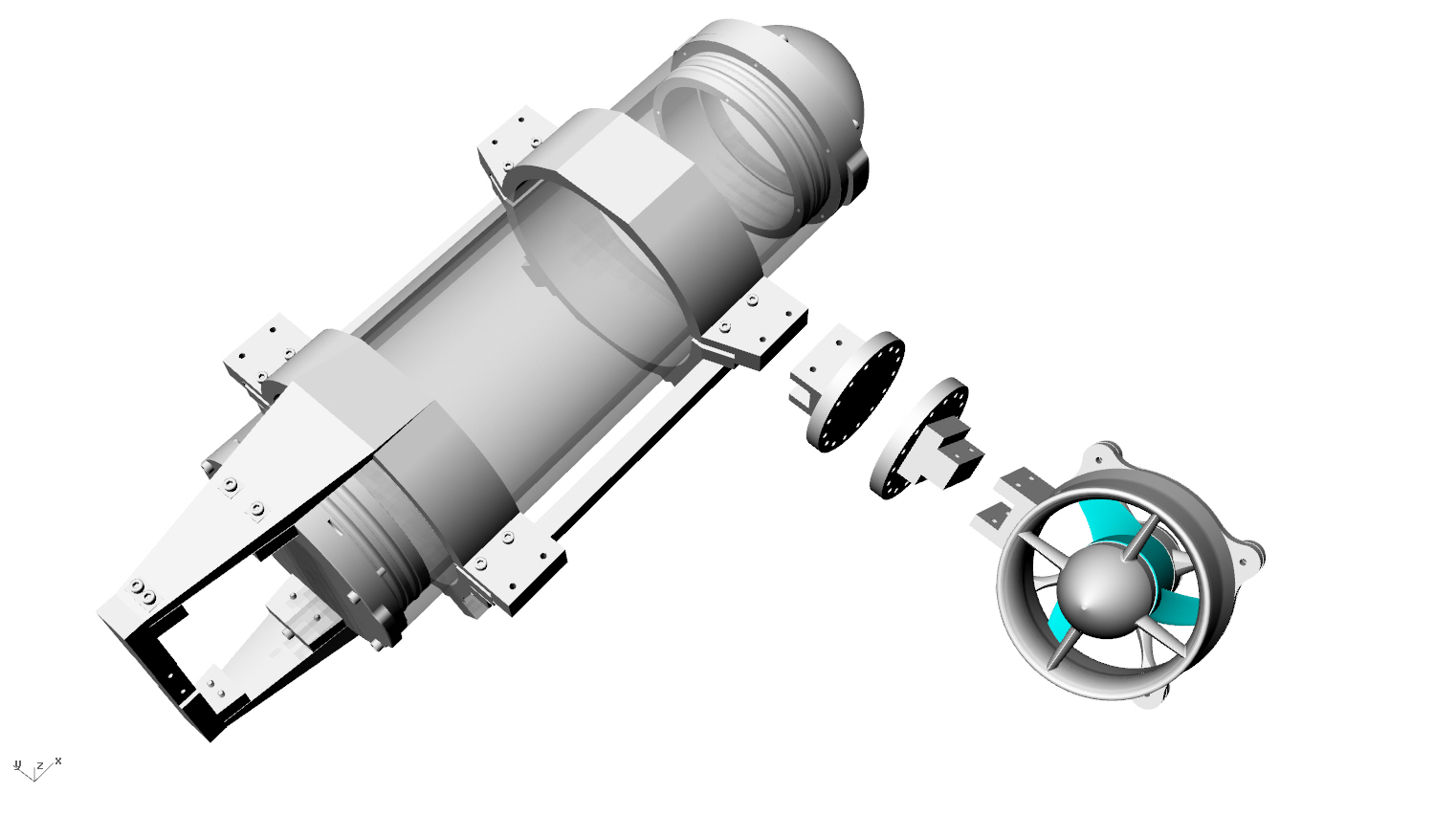}
  \caption{A rotation adapter provides five possible orientations within 90\degree.}
  \label{fig:mount4}
\end{figure}

As mentioned in the previous section, in low-cost ROVs characterised by fixed ballast, ascend and descend movements are commonly achieved using vertical thrusters. A conventional procedure is to make vehicles positively buyoant to ensure they will return to the surface, should any power failure occurs. We designed a particular cylindrical structure that can be filled with steel balls or syntactic foam, in order to allow the user to set the desired buyoancy to the drone. A ballast cylinder is divided into two compartments, should precise settings of the center of mass are needed.

\begin{figure}[htp!]
  \centering
  \includegraphics[scale=0.15]{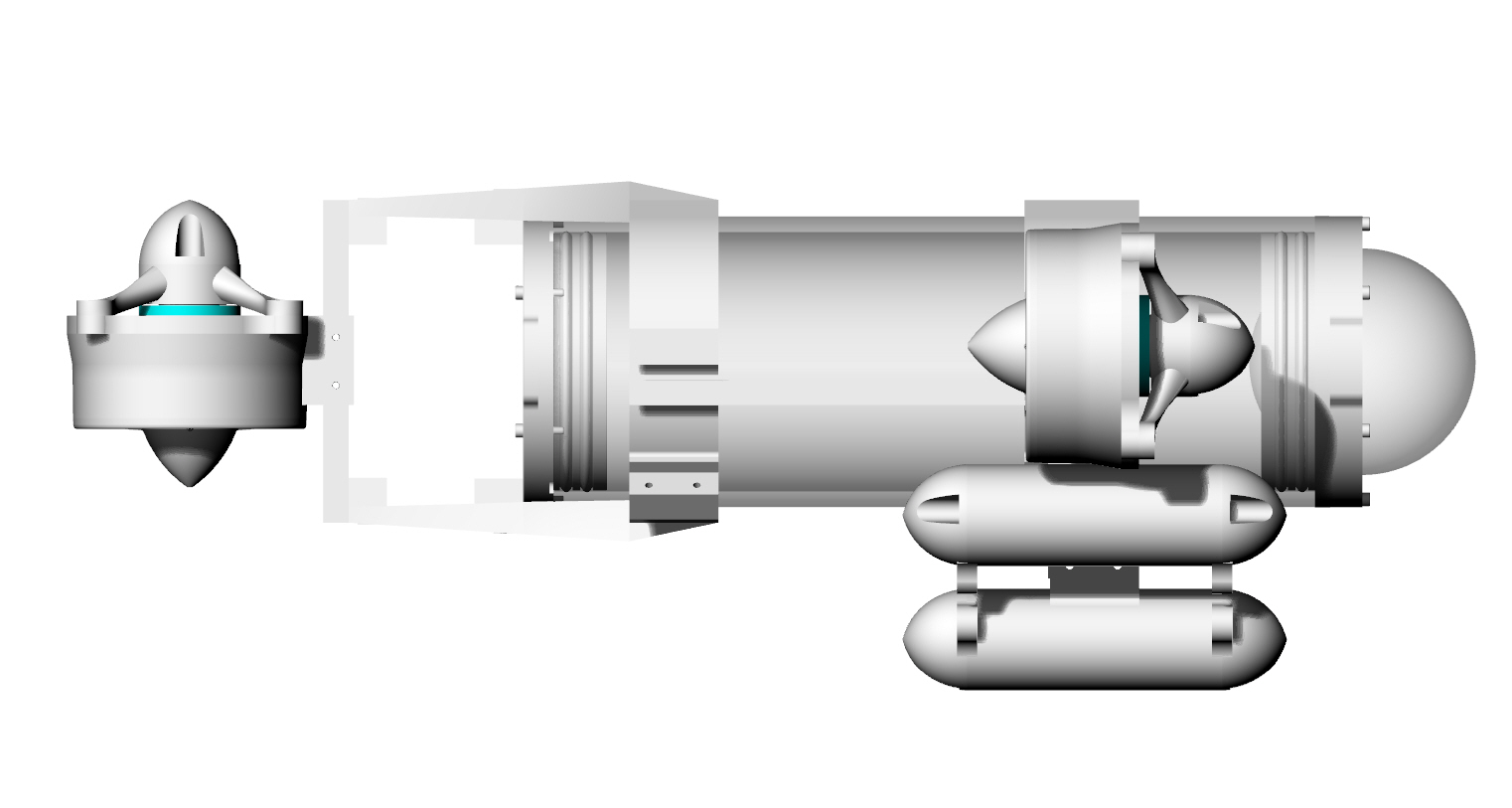}
  \caption{Horizontal view of a setup where four ballast cylinders are used mounting a split adapter onto the +135\degree mounting point and a split adapter onto the -135\degree mounting point of the enclosure clamp.}
  \label{fig:ballast1}
\end{figure}

\begin{figure}[htp!]
  \centering
  \includegraphics[scale=0.15]{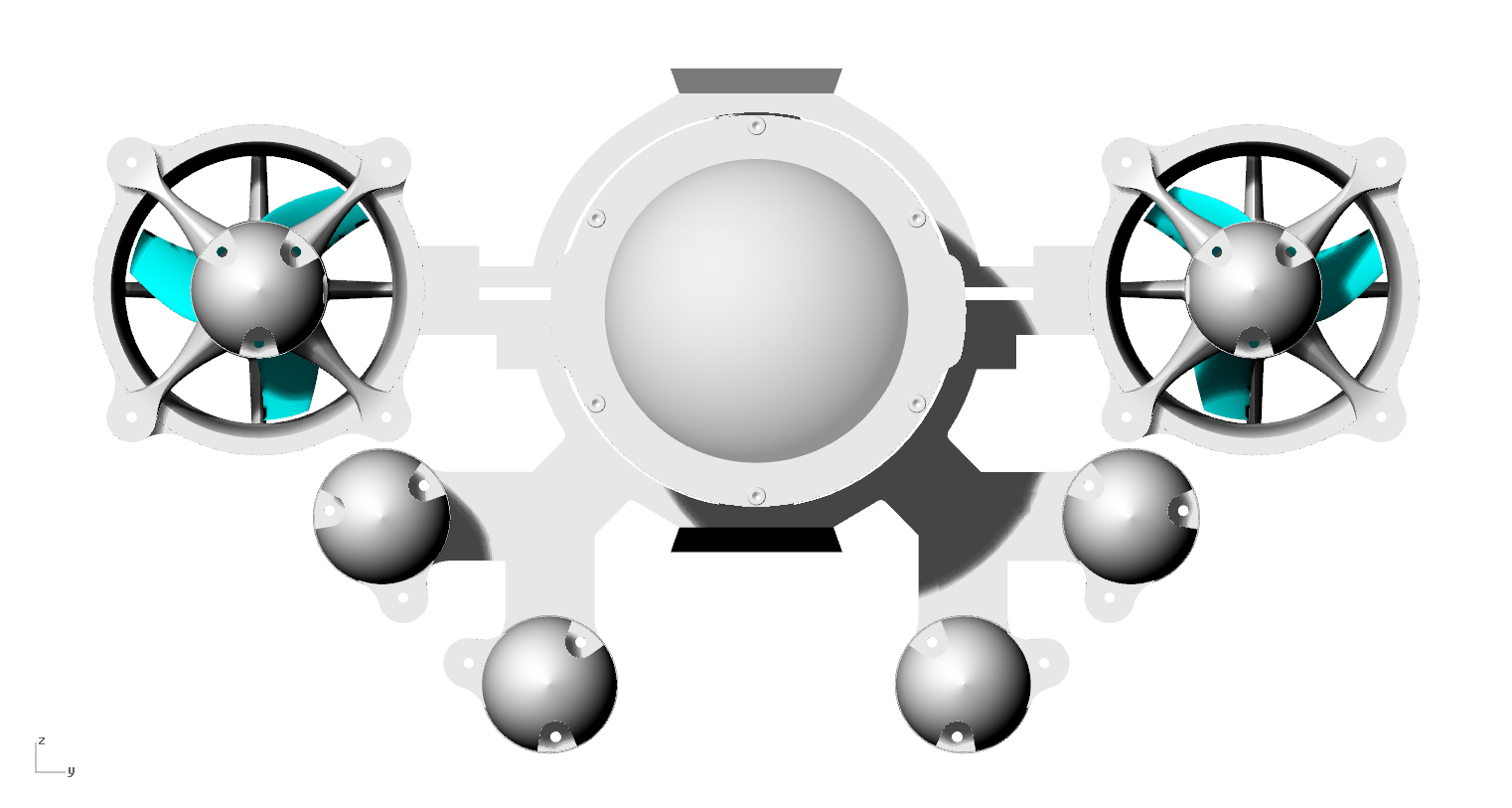}
  \caption{Frontal view of a setup where four ballast cylinders are used mounting two split adapters onto the +135\degree mounting point and two split adapters onto the -135\degree mounting point of the enclosure clamp.}
  \label{fig:ballast2}
\end{figure}

The ballast cylinder has the same mounting structure of the thurster, so that it can be mounted onto the same mounting points depicted above. Figure \ref{fig:ballast1} and \ref{fig:ballast2} show an example where four ballast cylinders are mounted onto the  +135\degree and -135\degree mounting points of the anterior enclosure clump.

The adapters, thrusters and ballasts can be mounted together using standard M3 nuts and M3-30mm bolts.

\begin{figure}[htp!]
  \centering
  \includegraphics[scale=0.22]{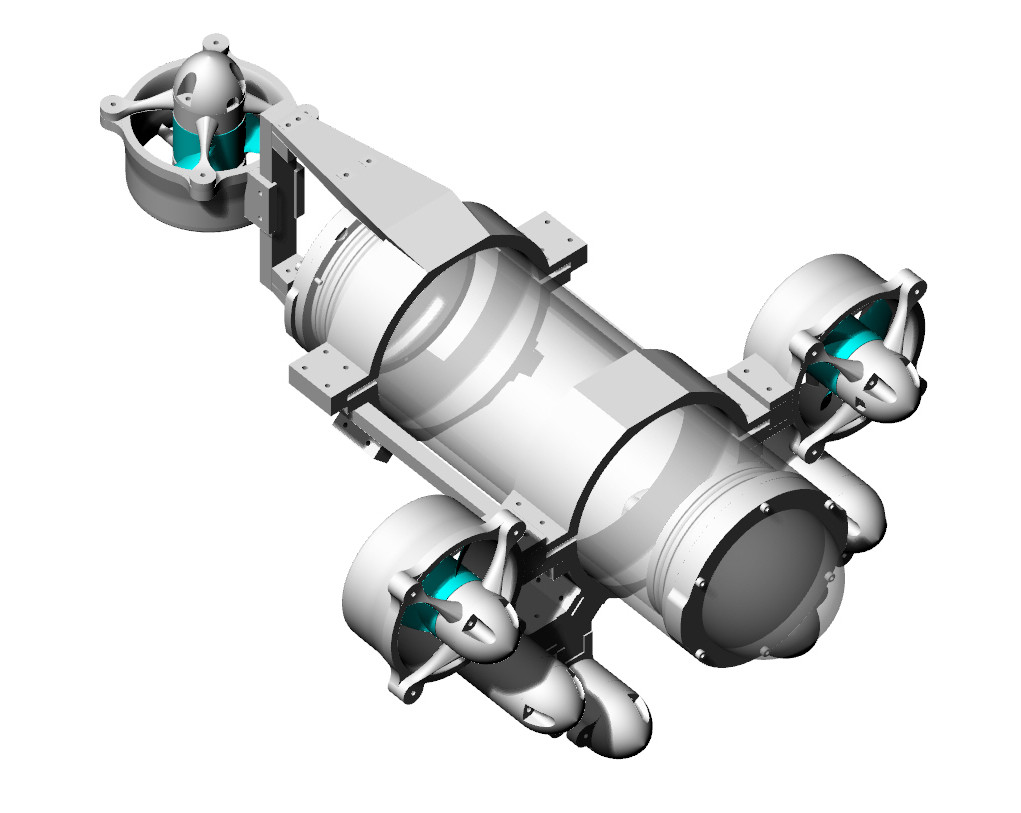}
  \caption{Underwater drone with a 3-thrusters frame configuration.}
  \label{fig:config1}
\end{figure}

The parts described above allow multiple frame configurations, each providing different maneuvering possibilites and degrees of freedom for the drone movements. 
Figure \ref{fig:config1} shows perhaps the simplest setup with a 3 thrusters frame configuration. Here, two horizontal thrusters allow for forward/backward movements and yaw rotations (left and right turns). A vertical thruster mounted onto the tail mounting point of the drone provides pitch rotation possibilities (nose up and nose down movements). As the only vertical thruster is placed far away from the center of mass of the robot, ascent and descend movements are not possible with such a configuration. 

\begin{figure}[htp!]
  \centering
  \includegraphics[scale=0.20]{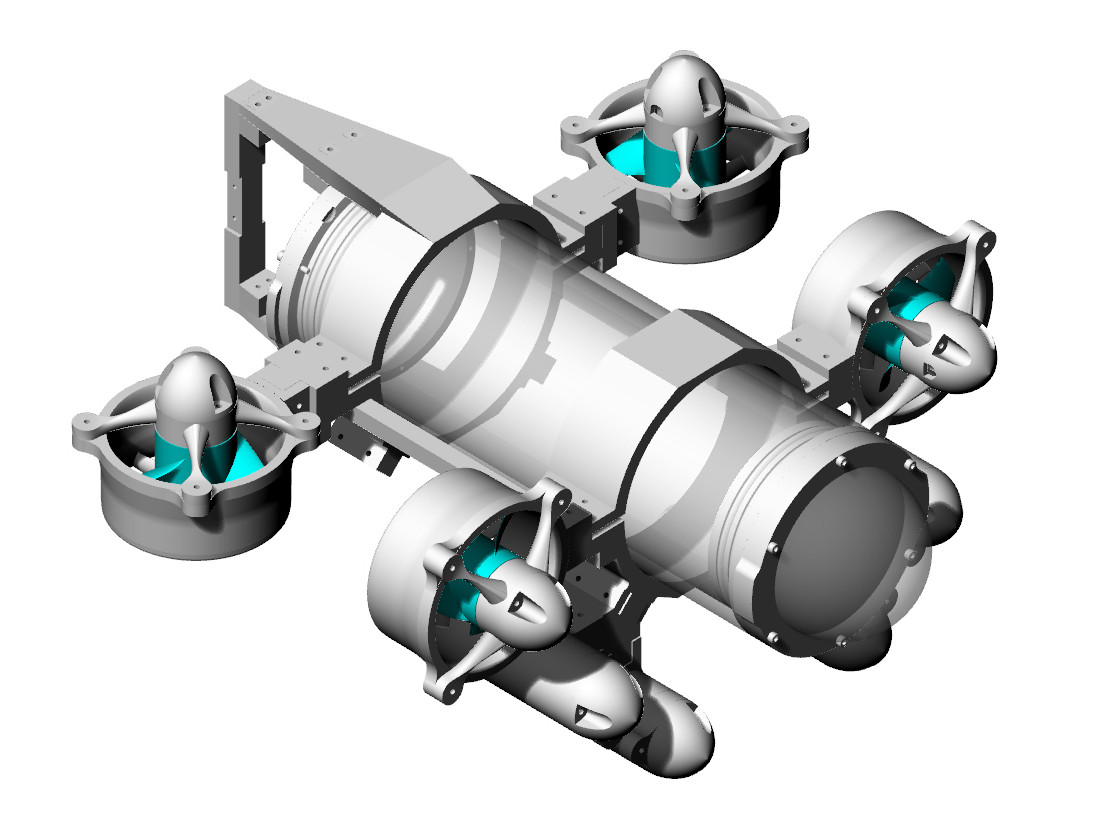}
  \caption{The drone with a 4-thrusters frame configuration. Vertical propellers are closer to the center of mass, allowing ascend and descent movements.}
  \label{fig:config2}
\end{figure}

\begin{figure}[htp!]
  \centering
  \includegraphics[scale=0.28]{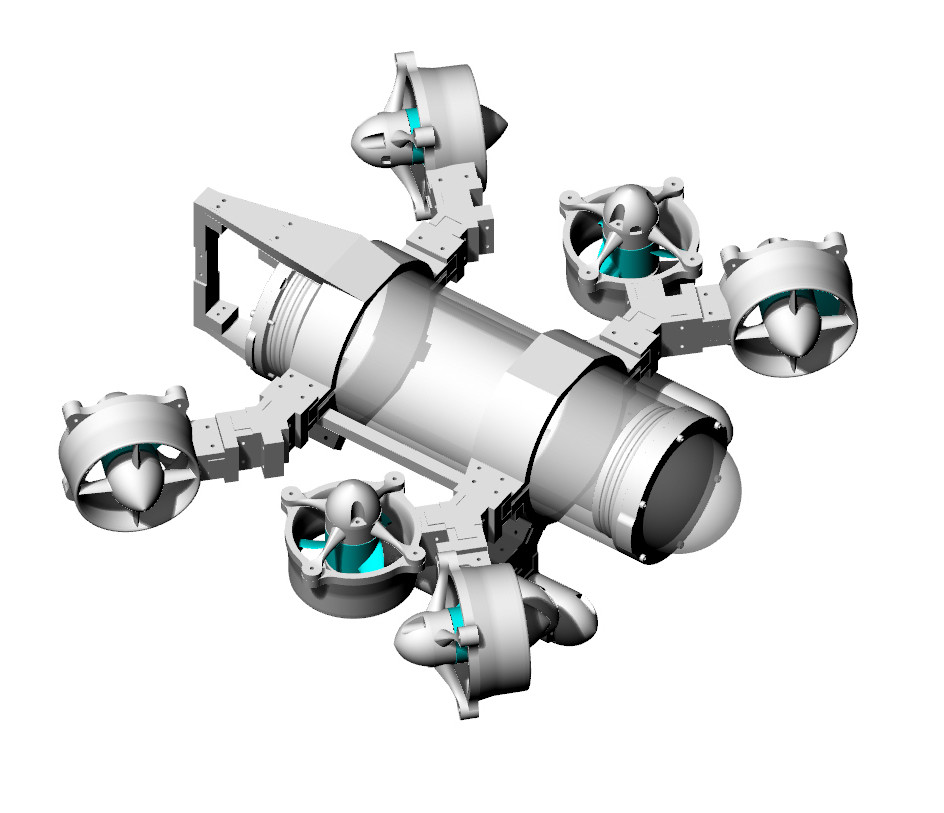}
  \caption{Vectored 6-thrusters frame configuration. This configuration allows for better control on the robot: forward, backward and lateral movements; ascend and descend movements; yaw and roll rotations.}
  \label{fig:config3}
\end{figure}

Although ballast weights can be mounted to adjust the position of the center of mass, if ascend and descend movements are required, additional vertical thrusters can be mounted onto the posterior enclosure clamp (see Figure \ref{fig:config2}) or onto a split adpater at the anterior enclosure clamp together with a 45\degree thruster (as in the vectored frame configuration depicted in Figure \ref{fig:config3}). This configuration may still require adjustments of the center of mass position through ballast, which can be plugged at the bottom of the posterior clamp or at the tail mounting point.

The vectored frame configuration depicted in Figure \ref{fig:config3} allows for full control on the robot: forward, backward and lateral movements; ascend and descend movements; yaw and roll rotations.

\section{A functional prototype}
\label{sec:proto}

\begin{figure}[htp!]
  \centering
  \includegraphics[scale=0.52]{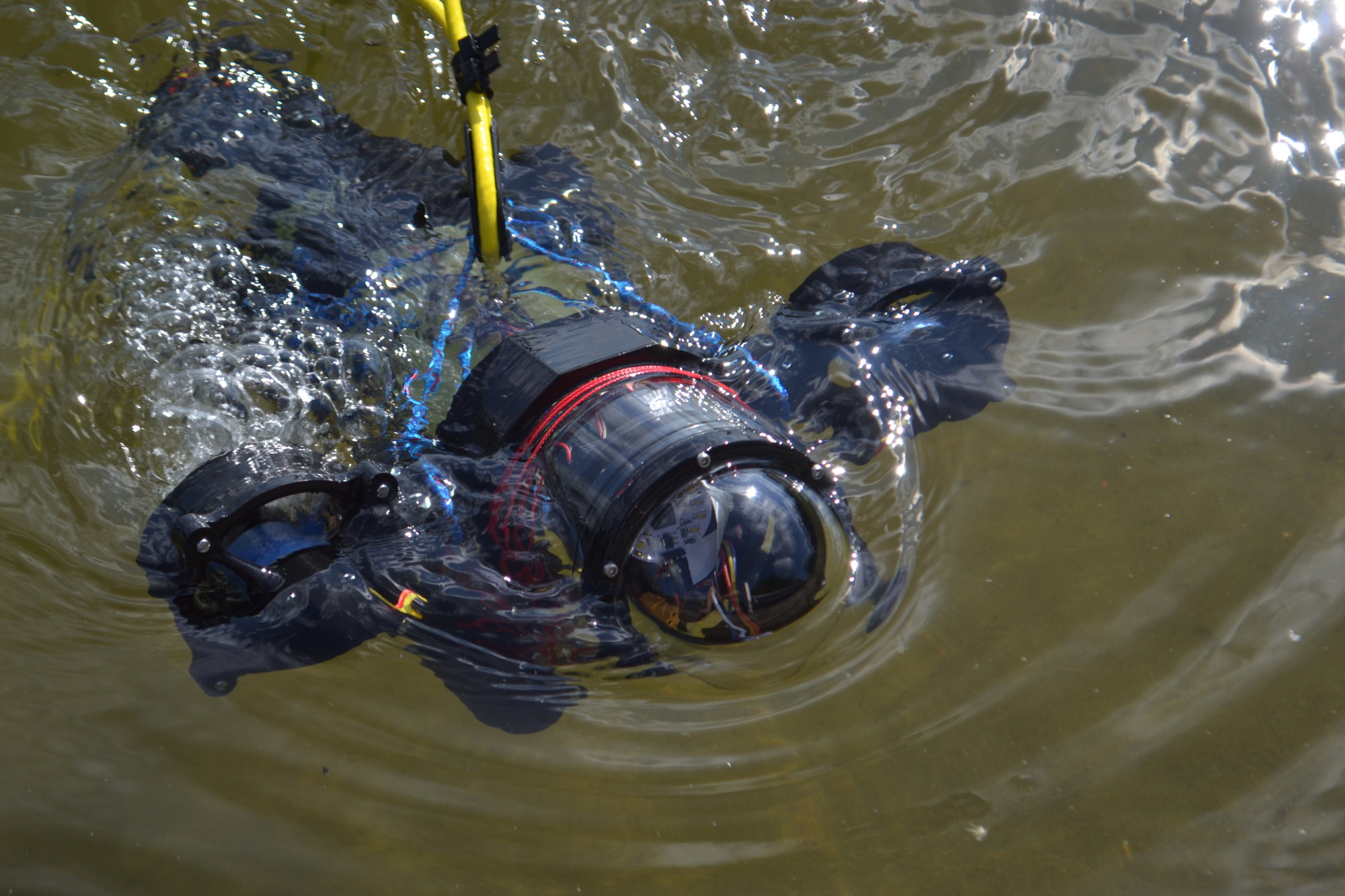}
  \caption{A picture of the drone during a dive in the Dahme river in K\"{o}penick, Berlin.}
  \label{fig:dive}
\end{figure}

In this section, we present an implementation of a functional prototype using a three thrusters frame configuration, as depicted in Figure \ref{fig:config1}. Section \ref{sec:hw} describes the electronics of the drone. In section \ref{sec:sw} we describe the software adopted for controlling the robot. We carried out several dives in different lakes and rivers around Berlin. Figure \ref{fig:dive} shows the drone emerging from a dive in the Dahme river in K\"{o}penick, Berlin.

\subsection{Hardware description}
\label{sec:hw}
We adopted low-cost and opensource, when available, solutions for implementing the electronics of the drone. In particular, we set up a Raspberry Pi 3 model B - equipped with a Quad Core 1.2GHz Broadcom BCM2837 64bit CPU and 1GB RAM - as a companion computer to an opensource flight controller, namely PX4 Pixhawk, sold by mRobotics. Pixhawk is an open-hardware project aiming at providing high-end autopilot hardware to the public at low costs \footnote{\url{https://pixhawk.org/}}. The board is equipped with several sensors, including a gyroscope, an accelerometer, a magnetometer and a barometer. 

We installed three Turnigy DST-700 brushless motors, each initially connected to the Pixhawk flight controller through a N-channel MOSFET Afro 12A Electronic Speed Controller, which provides PWM control of the motor speed. As we experienced a damage in a ESCs connected to a propeller that was put too much under torque stress, we opted for mounting Afro 20A ESCs instead of 12A ones.

A Turnigy 5000mAh 4S 25C Lipo Battery Pack is adopted to power the system. Power is distributed to the ESCs through a Matek power distribution board, to the Pixhawk through a mRobotics Power Module and to the Rasperry Pi through a 5V 3A Switching Power UBEC.

A Raspberry Pi V2.1, 8 MP 1080P camera module is installed on the Rapsberry Pi to stream HD videos to the surface computer. A Lisiparoi low-light led module is installed onboard. The drone is connected to a surface computer through a 25 meters neutrally buyoant RJ45 tether which provides data communication. The user can teleoperate the drone using a game controller. We tested a XBox360 controller and a ASUS TV500BG Bluetooth gamepad. Cheap cardboard VR goggles has been used to stream the video recorded from the Raspberry Pi camera onto a smartphone to ease the teleoperation of the drone. 

Electronics need to be put into a watertight container. We acquired a 4" watertight enclosure, a compatible aluminium end cap and a glass dome cap, and o-ring sealed flanges from the BlueRobotics online store. Motor power cables and tether are inserted into the watertight enclosure though cable penetrators and sealed through J-B Marine Weld marine epoxy. We acquired also a BlueRobotics waterproof switch that allows turning on/off the drone without the need of opening the watertight enclosure. The switch is connected upstream to the electronics. A pressure valve has been installed for vacuum seal tests and for simplifying opening/closing the watertight enclosure.

We acquired also a BlueRobotics Bar30 high resolution waterproof pressure and temperature sensor. The sensor comes in a watertight penetrator, thus it can be safely used and connected into the Pixhawk board.

\subsection{Software description}
\label{sec:sw}
We rely entirely on open-source software solutions for controlling the drone. In particular, we adopted Raspian as operating system for the Raspberry Pi \footnote{\url{https://www.raspberrypi.org/downloads/raspbian/}} and we installed ArduSub \footnote{\url{https://www.ardusub.com/}} onto the PX4 Pixhawk. ArduSub is a fully-featured open-source solution for remotely operated underwater vehicles, initially developed by BlueRobotics as derived from the ArduCopter code, now is part of the ArduPilot project. ArduSub has several pre-installed functionalities, including feedback stability control, depth and heading hold, and autonomous navigation using pre-coded trajectories. 

MAVlink is adopted as a communication protocol. QGroundControl is installed into the surface computer, tablet or smartphone for providing dive control and mission planning to the drone. QGroundControl, initially developed as the control station software for the Pixhawk Project, is an open-source application used at several research centers and by a growing communitiy \footnote{\url{http://qgroundcontrol.org/about}} for control of flying, terrestrial and underwater drones.

Finally, the open source SimonK firmware \footnote{\url{https://github.com/sim-/tgy}} has been compiled and deployed onto the Afro ESCs using an Afro USB programming tool, in order to allow forward/reverse rotations of the propellers and to match the parameters configured in ArduSub.

\section{Conclusions}
\label{sec:conclusions}
In this paper, we presented open source models for building low-cost underwater drones and for customising their thrusters and ballast configurations. The models are optimised for rapid prototyping through 3D printers. 

After introducing the different parts and adapters that can allow for multiple frame configurations and, thus, different grades of maneuverability, we presented a functional prototype. In particular, we built a three-propellers underwater drone based on open source hardware and software solutions. The prototype has been successfully tested several times in dives near Berlin. 

Although the reliability of the 3D printed parts have been tested only in relatively shallow waters (\textless 10 meters), we strongly believe that making them freely available online will motivate the general public to build and to experiment with underwater drones, and consequently speed up the development of innovative solutions and applications. 

Nonetheless, we are planning to carry out further tests in deeper waters and under different frame configurations. Moreover, we are currently exploring possibilities for  autonomous learning of robot control strategies.

\bibliographystyle{IEEEtran}
\bibliography{./paper.bib}
 
\end{document}